\documentclass[11pt,a4paper]{article}
\usepackage[hyphens]{url}
\usepackage{emnlp2020}
\usepackage[T1]{fontenc}
\usepackage{times}
\usepackage{latexsym}
\usepackage[utf8]{inputenc}
\usepackage[english]{babel}
\usepackage{soul}
\usepackage{makecell}
\usepackage{arydshln}
\usepackage{amsmath}
\usepackage{amssymb}
\usepackage{float}
\usepackage{tabularx}
\usepackage{tikz}
\usepackage{tikz-dependency}
\usepackage[inline]{enumitem}
\usepackage{caption}
\usepackage{subcaption}
\usepackage{multirow}
\usepackage{tabulary}
\usepackage{bm}
\usepackage{rotating}

\usetikzlibrary{matrix}

\newcommand{\secref}[1]{\S\ref{#1}}

\newcommand{\inputtok}{$I_{\mathrm{tok}}$~}
\newcommand{\outputtok}{$O_{\mathrm{tok}}$~}
\newcommand{\outputexp}{$O_{\mathrm{exp}}$~}

\usepackage{microtype}

\aclfinalcopy 

\title{Syntax-driven Iterative Expansion Language Models\\for Controllable Text Generation}

\author{Noe Casas\textsuperscript{\textdagger}\textsuperscript{\textasteriskcentered},~~~
Jose A. R. Fonollosa\textsuperscript{\textasteriskcentered},~~~
Marta R. Costa-jussà\textsuperscript{\textasteriskcentered} \\
\textsuperscript{\textdagger} Lucy Software, United Language Group \\
\textsuperscript{\textasteriskcentered} TALP Research Center, Universitat Politècnica de Catalunya \\
\texttt{\{noe.casas,jose.fonollosa,marta.ruiz\}@upc.edu} \\
}

\date{}

\begin{document}
\maketitle
\begin{abstract}
The dominant language modeling paradigm handles
text as a sequence of discrete tokens.
While that approach can capture the latent structure
of the text, it is inherently constrained to sequential
dynamics for text generation.
We propose a new paradigm for introducing a syntactic
inductive bias into neural text generation,
where the dependency parse tree
is used to drive the Transformer model to generate sentences
iteratively.

Our experiments show
that this paradigm is \mbox{effective} at text generation,
with quality between LSTMs and Transformers, and comparable diversity,
requiring less than half their decoding steps, and
its generation process allows direct control over
the syntactic constructions of the generated text, enabling the
induction of stylistic variations.
\end{abstract}

\section{Introduction}

The currently dominant text generation paradigm is based on
generating a sequence of discrete tokens in a left-to-right
autoregressive way.
Most neural language models (LMs) fall into this autoregressive
generation category.
Some neural architectures are sequential in nature, such as
those based on recurrent neural networks (RNNs),
lending themselves naturally to the autoregressive approach
when used together with teacher forcing \cite{williams1989teacherforcing}.
Other architectures, such as Transformer
\cite{vaswani2017attention}, while not intrinsically sequential,
have also been targeted for sequential generation.
On the other hand, some recent lines of research have focused
on nonsequential generation.
In this work, we propose a new paradigm for text generation
and language modeling called Iterative Expansion Language Model,
which generates
the final sequence following a token ordering defined by the
sentence dependency parse by iteratively expanding each
level of the tree.

\section{Related Work} \label{sec:related}

In this section, we provide an overview of works related to ours,
including
dependency tree-driven LMs (\secref{sec:deplms}),
syntax-driven generation (\secref{sec:relatedsyntax}),
insertion-based approaches  (\secref{sec:insertiongen})
and
iterative refinement approaches (\secref{sec:relatediterative}).

\subsection{Dependency LMs}\label{sec:deplms}

The use of dependency parse trees to drive a language model
was first proposed by \citet{chelba1997structure}, with
a similar structure to an $n$-gram LM, but where
the context of a word is its preceding bigram plus
a list of preceding words whose parent does not precede it.
\citet{shen2008new} make use of the dependency
tree in a probabilistic LM, computing the probability
of each word conditioned on its parent and the sibling words
between both.

\citet{mirowski-vlachos2015dependency} propose a dependency
LM based on RNNs, where the dependency tree
is decomposed into a collection of unrolls, that is, paths from
the root to one of the leaves, and where the probability of a word
can be predicted from these unrolls.
\citet{buys-blunsom2018neural} propose a shift-reduce
transition-based LSTM \cite{hochreiter1997long}
dependency LM that can be used for
language modeling and generation by means of
dynamic programming.

\subsection{Syntax-driven Generation} \label{sec:relatedsyntax}

Recurrent neural network grammars \cite{dyer2016rnng} are
recursive models that operate with a stack of symbols that
can be populated with terminals or nonterminals, or ``reduced''
to generate a syntactic constituent, obtaining as a result
a sentence and its associated constituency parse tree.

\citet{shen2018neural} use skip-connections to integrate
constituent relations with RNNs,
learning the underlying dependency structures
by leveraging a syntactic distance together with structured
attention.

\citet{akoury2019synst}
use a simplified constituency tree as latent variables,
modeling it autoregressively to later use it
as input for a non-autoregressive transformer that generates
the output sentence.

Ordered neurons \cite{shen2018ordered} are modified LSTMs
where the latent sentence tree structure
is used to control the dependencies between recurrent units
with a special ``master'' input and forget gates.

\subsection{Insertion-based Generation} \label{sec:insertiongen}

\citet{stern2019insertiontransformer} propose a conditional
generative model that iteratively generates tokens
plus the position at which they should be inserted within the sequence. 
\citet{emelianenko2019sequence} further propose to optimize
the generation order by sampling from the ordering
permutations. Instead, \citet{chan2019kermit} optimize a lower bound
of the marginalized probability over every possible ordering.

\citet{gu2019insertion} handle the generation order as a latent
variable that is captured as the relative position through self-attention,
optimizing the ELBO to train the model.

Levenshtein Transformer \citep{gu2019levenshtein} is a
non-autoregressive approach
trained with reinforcement learning (RL) to generate token
insertion and deletion actions. While it benefits from the same 
generation speed-ups over autoregressive models as our model,
it has the added  difficulty of learning an insertion/deletion 
policy using RL without any linguistically or 
empirically motivated priors, which can be slow or difficult to 
obtain convergence in practice. By comparison, our approachmakes 
uses a linguistically motivated prior for word insertion in a
fully supervised way, avoiding the optimization difficulties of RL.

\citet{welleck2019nonmonotonicsequential} use cost minimization
imitation learning to learn a policy to generate a binary
tree that is used to drive the token generation.

\subsection{Iterative Refinement} \label{sec:relatediterative}

\citet{lee2018iterativenar} propose a latent variable non-autoregressive
machine translation model where first the target length is
predicted by the model, and then, the decoder is iteratively applied
to its own output to refine it.

Mask-predict \cite{ghazvininejad2019MaskPredict} also predicts
the target sentence length and then non-autoregressively predicts
the sentence itself, iteratively refining it a fixed number
of times, masking out and regenerating the tokens it is least
confident about.
\citet{lawrence2019attending} follow a similar
approach and start with a sequence of placeholder tokens
(all the same) of a specified length,
and they iteratively replace them
with normal tokens via masked LM-style inference. As the masking strategy
for the training data, the authors propose different stochastic
processes to randomly select which placeholders are to be uncovered.

\section{Iterative Expansion LMs} \label{sec:approach}

Our proposal is to train a new kind of language model where the
token generation order is driven by the dependency
parse tree of the sentence and where the generation process
is iterative.

\begin{figure}[ht]
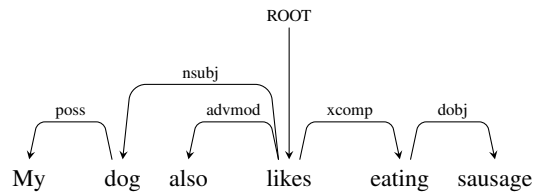

\begin{center}
\begin{dependency}[font=\footnotesize,text only label, label style={above}]
\begin{deptext}[column sep=0.2cm]
My \&[.4cm] dog \& also \&[.4cm] likes \&[.4cm] eating \& sausage \\
\end{deptext}
\depedge{2}{1}{poss}
\depedge{4}{2}{nsubj}
\depedge{4}{3}{advmod}
\depedge{4}{5}{xcomp}
\depedge{5}{6}{dobj}
\deproot{4}{ROOT}
\end{dependency}
\end{center}
\caption{Example of dependency parse tree.}
\label{fig:deptreeexample}
\end{figure}

The input vocabulary contains terminal tokens as well as
non-terminal special tokens called dependency placeholders,
each of which is associated with
one of the possible dependency relations to the heads. For the dependency
tree in Figure \ref{fig:deptreeexample}, the dependency
placeholders are
\texttt{[poss]},
\texttt{[nsubj]},
\texttt{[advmod]},
\texttt{[xcomp]},
\texttt{[dobj]} and
\texttt{[ROOT]}.

The input of the first iteration is the sequence with the
\texttt{[ROOT]} element.
At each iteration, the model receives as input a sequence \inputtok
with tokens from the input vocabulary
and non-autoregressively generates
two new sequences, each with the same length as the input.

The first output sequence, \outputtok, contains tokens from a vocabulary
with all possible textual tokens (terminal tokens).
The second output, \outputexp, is a sequence of tokens called
expansion placeholders, which are taken from a separate
vocabulary.
Each expansion placeholder is associated
with a pattern describing the left and right dependencies of the token
at that position in the \outputtok sequence. An example
of dependency expansion could be \mbox{\texttt{[nsubj-advmod-HEAD-xcomp]}}
for the word ``likes'' in the dependency parse tree from Figure
\ref{fig:deptreeexample}.

After each iteration, the output of the model is expanded.%
\footnote{The expansion of the output to be fed as input in the
next iteration occurs in the CPU outside of the neural model itself.} This
consists of creating a new sequence by combining the tokens
from \inputtok, \outputtok and \outputexp.
This process is illustrated in Figure \ref{fig:expansionexample},
making use of the dependency tree from Figure \ref{fig:deptreeexample}.

When there is a padding token \texttt{[pad]} in the output
(either \outputtok or \outputexp),
this means that the output at that position is ignored when
computing the loss function. This occurs when the terminal
token has already been computed in previous iterations and
has therefore been received as part of \inputtok,
and the model does not need to compute it again.

Note also that an empty dependencies token \texttt{[HEAD]} marks
the end of a branch and that there is no need for an end of sequence
token \texttt{<eos>}.
As shown in the example from Figure \ref{fig:deptreeexample},
the generation of different branches occurs in
parallel, needing only 3 iterations to generate a 6-token sentence.

\begin{figure}[ht]
\fontsize{8pt}{12pt}\selectfont
\setlength{\tabcolsep}{3pt}
\begin{tabularx}{\columnwidth}{p{6mm} c }
\multicolumn{2}{l}{Iteration 1}\\
\hline
\inputtok: & \texttt{[ROOT]}\\
\outputtok: & likes\\
\outputexp: & \texttt{[nsubj-advmod-HEAD-xcomp]}\\
\end{tabularx}
\begin{tabularx}{\columnwidth}{p{6mm} c c c c}
\multicolumn{2}{l}{Iteration 2\hfill}\\
\hline
\inputtok: & \texttt{[nsubj]} & \texttt{[advmod]} & likes & \texttt{[xcomp]} \\
\outputtok: & dog & also & \texttt{[pad]} & eating \\
\outputexp: & \texttt{[poss-HEAD]} & \texttt{[HEAD]} & \texttt{[pad]} & \texttt{[HEAD-dobj]} \\
\end{tabularx}
\begin{tabularx}{\columnwidth}{p{6mm} c c c c c c}
\multicolumn{2}{l}{Iteration 3}\\
\hline
\inputtok: & \texttt{[poss]} & dog & also & likes & eating & \texttt{[dobj]} \\
\outputtok: & my & \texttt{[pad]} & \texttt{[pad]} & \texttt{[pad]} & \texttt{[pad]} & sausage \\
\outputexp: & \texttt{[HEAD]} & \texttt{[pad]} & \texttt{[pad]} & \texttt{[pad]} & \texttt{[pad]} & \texttt{[HEAD]} \\
\end{tabularx}
\caption{Example of iterative text generation.}
\label{fig:expansionexample}
\end{figure}

The strategy for composing tree expansion tokens
(e.g., \texttt{[nsubj-advmod-HEAD-xcomp]}) may
not scale well when single words have many direct
dependencies. To alleviate this, we introduce
a preprocessing step to modify the dependency tree so that
every word has at most one dependency to the left and one
to the right. For each word with more than one dependency
on any of its sides, we rearrange the tree to force left-to-right dependencies.
Although this \textbf{tree binarization} reduces the degree of parallelism,
it reduces data sparsity and allows handling constructions with a number of 
dependencies may otherwise be too large
for the model to properly capture, such as enumerations (e.g.,
``I bought a pair of shoes, an umbrella,
a beautiful jacket and a bracelet'').

Iterative expansion LMs can be naturally extended to subword
vocabularies, like byte-pair encoding \cite[BPE;][]{sennrich2016bpe}:
for each word, we decompose its node in the tree into as many nodes
as subwords in the word, rearranging the tree so that
the head of the old word is now the head of the first subword,
and each subsequent subword depends on the previous one, while every
dependency of the old word node now depends on the last subword.

\subsection{Neural Architecture} \label{sec:architecture}

The neural architecture proposed is based on a Transformer decoder
\cite{vaswani2017attention}.
To generate the dual output (terminal
tokens and expansion placeholders) we 
condition the generation of terminals on the expansions:
the probability
distribution over the expansion token space is generated first
by projecting from one of the intermediate layers' hidden
states. We sample from it and use the resulting expansion IDs as
an index to a trainable expansion embedding layer; the
embedded vectors are added to the hidden state
used to generate them for use as input to subsequent layers.

As described in Section \ref{sec:approach}, the input and output token
vocabularies are different: the latter only contains terminal
tokens (plus some special tokens such as \texttt{[PAD]}); the former
also contains dependency placeholders. However, for practical
purposes, at the model level, we define both vocabularies to be the
same, both with terminal tokens and dependency placeholders,
and we mask the entries of dependency placeholders in
the final softmax.

To inject the syntactic dependency information as input
into the model, we add a layer of learned positional embeddings
containing the position of the head of each token, and we refer to
this embedding layer as head position embedding.

The self-attention mask used in Transformer to force causality
is not used in our proposal.
The input is therefore not masked at all, and the
token predictions have access to the full input sequence.

\subsection{Training}

For training iterative expansion LMs, the main input of the model
is the tokens at one of the levels of the dependency parse tree
(\inputtok),
while the output is the following level tokens
(\outputtok) and
expansion placeholders (\outputexp).
A secondary input to the model are the dependency
indexes, which are used in the head position embedding.

The model is trained with the categorical
cross-entropy for both tokens
and expansion placeholders, then adding both sublosses
into the final loss (with equal weights). Tokens generated in previous
iterations appear as \texttt{[PAD]} tokens in the expected
output and are ignored when computing the loss.

Training takes place in batches; as the
trainable unit is a level transition, 
a training batch is composed of level transitions
from different sentences.

\subsection{Inference and Text Generation} \label{sec:inference}

In iterative expansion LMs, inference takes place iteratively.
The initial state is a batch of \texttt{[ROOT]} tokens, together
with the head positions initialized to the special value representing
the root node and, in constrained attention variants,
a mask with the self-dependency of the single node in each sentence
in the batch.
At each iteration, the model generates the probability distributions
for terminal tokens and expansion tokens.
We use nucleus sampling \citep{holtzman2019curious} to sample
from them. The terminal token sequences are expanded according to
the expansion tokens (see \secref{sec:approach}),
and these are the inputs for the following
iteration if there are still unfinished branches.
Before sampling from the token and expansion probability distributions,
we mask the \texttt{<unk>} token and the dependency placeholders to
avoid generating them.

Although iterative expansion LMs could be subject to beam search
across iterations, we have not covered such a possibility as
part of this work.

\section{Experimental Setup} \label{sec:expsetup}

\subsection{Unconditional Text Generation} \label{sec:textgen}

We conducted experiments on unconditional text generation following
the methodology used by \citet{caccia2018language}. The goal
is to assess both the quality and diversity of the text
generated by the model and the baselines.
For the quality evaluation, we use the BLEU score \cite{papineni2002bleu}
over the test set, where each generated sentence is evaluated against the
whole test set as a reference.
For diversity, we used the self-BLEU score \cite{zhu2018texygen},
computed using as references the rest of the generated sentences.
For each model,
the temperature of the final softmax $\tau$ is tuned to generate
text in the closest quality/diversity regime to the training data.

Iterative expansion LMs are compared against a standard
LM baselines,
namely, AWD-LSTM\footnote{Abbreviation of ASGD weight-dropped LSTM,
where ASGD stands for averaged stochastic gradient descent.}
\cite{merity2018regularizing}
and a Transformer LM \cite{vaswani2017attention},
both with \mbox{word (w)} and BPE subword (sw) vocabularies.
The models were trained on the EMNLP2017 News dataset, which
contains news in English,
enriched with dependency annotations by \texttt{corenlp},
an automatic annotation tool that provides pre-trained models.
Syntax-driven generation baseline models were not included
because the only model with an available implementation
that is able to do unsupervised text generation
are RNNGs, but they proved not to scale even to medium-sized
datasets like EMNLP2017 News.
When sampling from models,
we use nucleus sampling \cite{holtzman2019curious},
a form of ancestral sampling that constrains the candidate pool by
discarding the distribution tail.
Samples from the training and validation data are
included for reference.

Full hyperparameters and
data processing details are described
in Appendices \ref{sec:hyperparams} and \ref{sec:preprocessing}.

\subsection{Style Variation} \label{sec:style}

Iterative expansion LMs drive the generation of text
with the dependency parse tree. It is possible to influence the
generated trees by altering artificially the probability of the different
expansion tokens. To demonstrate this, we modified the
decoding process of iterative expansion LMs to 
force the probability of generating adjectival
constructions to be higher than normal,
aiming at generating a more descriptive style:
during decoding, we multiply the probabilities
of the expansion placeholders that express adjectival
dependencies (i.e. those containing adjectival modifier ``amod''
relations), and renormalize the probabilities
by dividing by the sum.

We conducted this experiment with the word-level models trained
on EMNLP2017 News data. We compute the ratio of adjectives per sentence
to verify the increased presence of adjectives,
while controlling quality and diversity measures over the generated text
for potential degradation.

\section{Results and Analysis} \label{sec:results}

We assess the ability of iterative expansion LMs to unconditionally
generate text in terms quality (BLEU-5) vs. diversity (self BLEU-5), 
comparing against sequential baselines, each with a softmax
temperature $\tau$ tuned separately.

\begin{figure*}[ht]
\centering
\includegraphics[width=\linewidth]{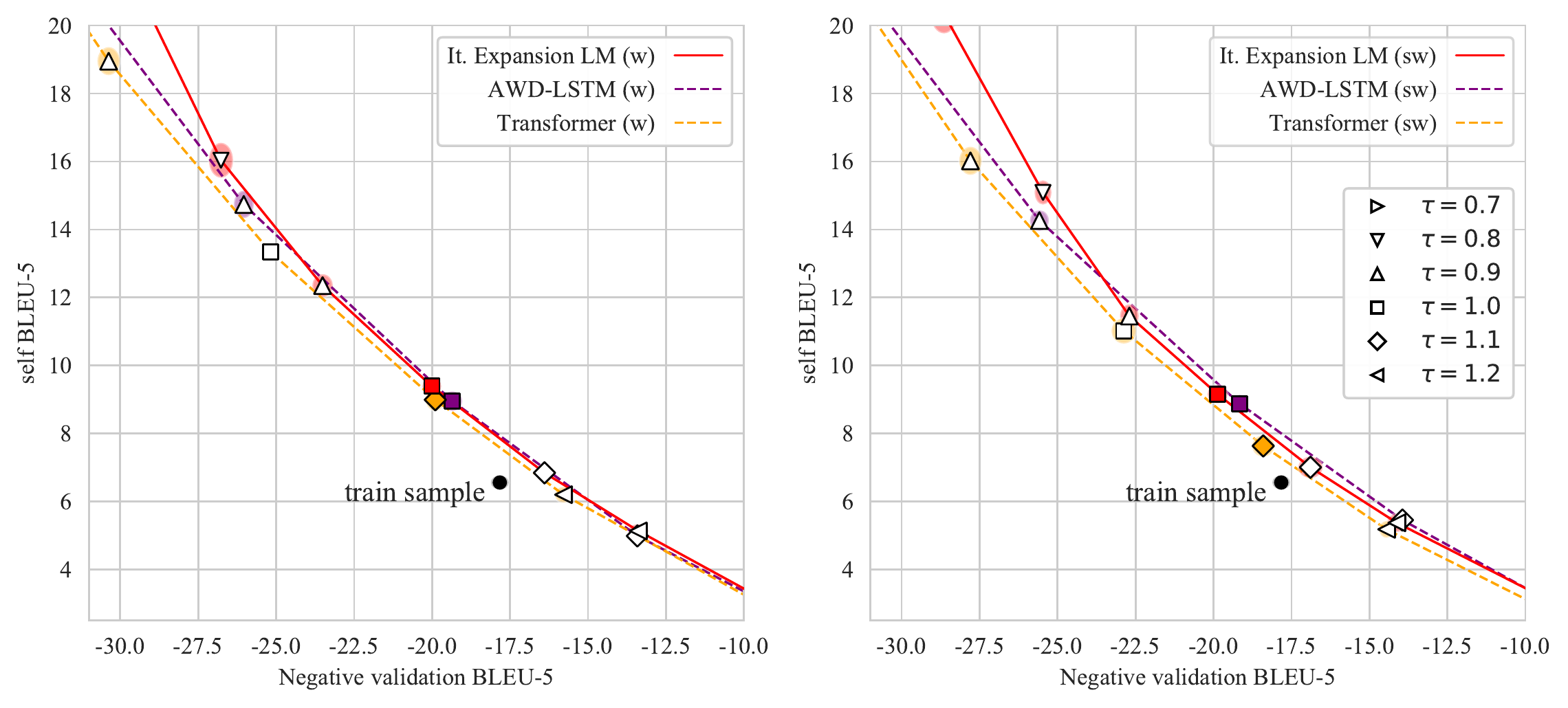}
\caption{Quality vs. diversity on EMNLP2017 News (BLEU-5). Models
with \textbf{word-level vocabulary on the left} and
\textbf{subword-level on the right}.
The point marker is color-filled for the chosen
value of $\tau$. Each point represents the average over 20
generated text samples, and is surrounded by a small colored ellipse
representing the standard deviation.
\label{fig:bleuvsselfbleu5}}
\end{figure*}

\begin{table*}[ht!]
\fontsize{10pt}{14pt}\selectfont
\centering
\begin{tabular}{c|rr|rr|rr}
\multirow{2}{*}{$\tau$}
&  \multicolumn{2}{c}{\textbf{\textsc{ItExp} (w)}} 
&  \multicolumn{2}{c}{\textbf{AWD-LSTM (w)}}
&  \multicolumn{2}{c}{\textbf{Transformer (w)}} \\
& \multicolumn{1}{c}{\textbf{valid} $\uparrow$} & \multicolumn{1}{c}{\textbf{self} $\downarrow$} & 
\multicolumn{1}{c}{\textbf{valid} $\uparrow$} & \multicolumn{1}{c}{\textbf{self} $\downarrow$} & 
\multicolumn{1}{c}{\textbf{valid} $\uparrow$} & \multicolumn{1}{c}{\textbf{self} $\downarrow$} \\
\hline

   0.70 & $30.1 \pm 0.8$                   & $22.3 \pm 1.0$                   & $39.2 \pm 0.9$           & $33.4 \pm 1.1$           & $40.5 \pm 0.6$              & $35.0 \pm 1.1$              \\
   0.80 & $26.8 \pm 0.8$                   & $16.0 \pm 1.0$                   & $33.0 \pm 0.7$           & $23.2 \pm 1.0$           & $35.8 \pm 0.7$              & $26.3 \pm 0.8$              \\
   0.90 & $23.5 \pm 0.7$                   & $12.4 \pm 0.7$                   & $26.0 \pm 0.6$           & $14.7 \pm 0.8$           & $30.4 \pm 0.7$              & $19.0 \pm 0.8$              \\
   1.00 & $\boldsymbol{20.0 \pm 0.6}$                   & $\boldsymbol{9.4 \pm 0.5}$                    & $\boldsymbol{19.4 \pm 0.6}$           & $\boldsymbol{9.0 \pm 0.6}$            & $25.2 \pm 0.5$              & $13.3 \pm 0.5$              \\
   1.10 & $16.4 \pm 0.5$                   & $6.8 \pm 0.5$                    & $13.4 \pm 0.4$           & $5.0 \pm 0.4$            & $\boldsymbol{19.9 \pm 0.6}$              & $\boldsymbol{9.0 \pm 0.6}$               \\
   1.20 & $13.4 \pm 0.6$                   & $5.1 \pm 0.4$                    & $9.0 \pm 0.5$            & $2.9 \pm 0.3$            & $15.8 \pm 0.5$              & $6.2 \pm 0.5$               \\
\end{tabular}\\
\vspace{5mm}
\begin{tabular}{c|rr|rr|rr}
\multirow{2}{*}{$\tau$}
&  \multicolumn{2}{c}{\textbf{\textsc{ItExp} (sw)}} 
&  \multicolumn{2}{c}{\textbf{AWD-LSTM (sw)}}
&  \multicolumn{2}{c}{\textbf{Transformer (sw)}} \\
& \multicolumn{1}{c}{\textbf{valid} $\uparrow$} & \multicolumn{1}{c}{\textbf{self} $\downarrow$} 
& \multicolumn{1}{c}{\textbf{valid} $\uparrow$} & \multicolumn{1}{c}{\textbf{self} $\downarrow$} 
& \multicolumn{1}{c}{\textbf{valid} $\uparrow$} & \multicolumn{1}{c}{\textbf{self} $\downarrow$} \\
\hline

   0.70 & $28.6 \pm 0.9$                    & $20.3 \pm 1.1$                    & $39.0 \pm 0.8$            & $33.5 \pm 1.1$            & $36.9 \pm 0.7$               & $30.6 \pm 1.2$               \\
   0.80 & $25.5 \pm 0.5$                    & $15.1 \pm 0.7$                    & $32.3 \pm 0.7$            & $22.4 \pm 0.7$            & $32.5 \pm 0.7$               & $22.4 \pm 1.0$               \\
   0.90 & $22.7 \pm 0.6$                    & $11.5 \pm 0.7$                    & $25.6 \pm 0.6$            & $14.3 \pm 0.6$            & $27.8 \pm 0.7$               & $16.0 \pm 0.8$               \\
   1.00 & $\boldsymbol{19.9 \pm 0.6}$                    & $\boldsymbol{9.2 \pm 0.5}$                     & $\boldsymbol{19.2 \pm 0.5}$            & $\boldsymbol{8.9 \pm 0.5}$             & $22.9 \pm 0.8$               & $11.0 \pm 0.7$               \\
   1.10 & $16.9 \pm 0.8$                    & $7.0 \pm 0.6$                     & $13.9 \pm 0.5$            & $5.5 \pm 0.4$             & $\boldsymbol{18.4 \pm 0.7}$               & $\boldsymbol{7.6 \pm 0.6}$                \\
   1.20 & $14.1 \pm 0.6$                    & $5.4 \pm 0.5$                     & $9.7 \pm 0.4$             & $3.3 \pm 0.3$             & $14.5 \pm 0.5$               & $5.2 \pm 0.5$                \\
\end{tabular}
\caption{Validation and self BLEU-5 scores of the text generated by
the \textbf{word-level (top)} and \textbf{subword-level (bottom)} models under study at different
temperatures $\tau$, showing the average and standard deviation
over 20 different generated text samples.
The selected generation regime is highlighted for
each model, being the closest to the training sample, which has
a validation BLEU-5 of 17.8 and a self BLEU-5 of 6.6.
}
\label{tab:tautuning}
\end{table*}

In order to tune the output softmax termperature $\tau$,
we generated text with each model at different temperatures
and chose the value of $\tau$ that was the most similar
to a sample from the training data in terms of 
BLEU-5 against a sample from the validation set (proxy for quality)
and self BLEU-5 (proxy for diversity).
Each model was used to generate 20 samples of 400 sentences,
and self-BLEU5 and validation-BLEU5 were computed over each of
them, taking the average and the standard deviation.
Figure \ref{fig:bleuvsselfbleu5} and
Table \ref{tab:tautuning} show these BLEU values,
highlighting the chosen $\tau$ for each model.
Given the low values for the standard deviation, we
decided not to include it in subsequent tables
to avoid unnecessary clutter.
Note that in all BLEU vs. self-BLEU figures, each model is
shown as a different line (each with its own color and/or
dashed pattern) and that the data points computed for each
temperature value are plotted with a specific marker shape
(square, diamond, triangle, or flipped triangle).

\begin{table*}[ht]
\centering
\fontsize{10pt}{14pt}\selectfont

\begin{tabular}{r c c c c c c c}
 & \multirow{2}{*}{$\tau$} & \textbf{Test BLEU-5}  & \textbf{Self BLEU-5}  & \textbf{AWD-LSTM} & \textbf{Transformer} & \textbf{GPT-2} \\
 & & \textbf{(quality $\uparrow$)} & \textbf{(diversity $\downarrow$)} & \textbf{perplex.} $\downarrow$ & \textbf{perplex.} $\downarrow$ & \textbf{perplex.} $\downarrow$ \\
\hline
AWD-LSTM (w) & $1.0$ & 22.9 & 8.9 & 37.0 & 47.9 & 99.5 \\
Transformer (w) & $1.1$  & 23.8 & 9.0 & 33.6 & 18.6 & 66.5 \\
\textsc{ItExp} (w) & $1.0$ & 23.7 & 9.4 & 40.8 & 40.7 & 85.2 \\
\hline
AWD-LSTM (sw) & $1.0$ & 22.7 & 8.9 & 43.5 & 56.9 & 113.5 \\
Transformer (sw) & $1.1$ & 22.1 & 7.6 & 37.5 & 31.6 & 77.1 \\
\textsc{ItExp} (sw) & $1.0$ & 23.6 & 9.2 & 45.2 & 49.2 & 97.1 \\
\hline
Train sample & - & 21.5 & 6.6 & 49.5 & 29.1 & 37.7 \\
Valid sample & - & 21.2 & 7.2 & 53.3 & 44.7 & 36.7 \\
\end{tabular}
\caption{\label{tab:quality} Quality and diversity on EMNLP2017, with $\tau$
generating the closest text to the validation data. }
\end{table*}

Apart from BLEU scores, we also include extra quality measures,
namely the perplexity obtained
by other language models:
an AWD-LSTM word-level LM and a Transformer word-level LM,
both trained on EMNLP2017 News, plus
\mbox{OpenAI} GPT-2 (1.5 B parameters) \cite{radford2019language}.
The results are shown in Table \ref{tab:quality}.

These results show how the generated text improves over AWD-LSTM
in terms of quality by all measures, with a comparable level
of diversity. In comparison to the Transformer, while the quality
measured with BLEU-5 is better for \textsc{ItExp}, the rest of the
quality measures indicate that the text generated by the Transformer
is of better quality.

\begin{table}[ht]
\centering
\fontsize{10pt}{14pt}\selectfont
\setlength{\tabcolsep}{3.5pt}
\begin{tabular}{c | c c c c }
\textbf{Adjective}  
 & \textbf{Adjs. per}
 & \textbf{Test} & \textbf{Self}\\  
\textbf{probability} & \textbf{sentence} & \textbf{BLEU-5} & \textbf{BLEU-5} \\
\hline
$\times 1$  & 1.2 & 23.7 & 9.4 \\
$\times 10$ & 3.4 & 21.3 & 8.4 \\
$\times 20$ & 4.2 & 20.6 & 8.8 \\
$\times 50$ & 5.2 & 19.8 & 8.9 \\
\end{tabular}
\caption{\label{tab:style} {\fontsize{10pt}{12pt}\selectfont \textsc{ItExp} (w, $\tau=1.0$)}
with increased adjectives.}
\end{table}

The results of the styled text generation experiments, shown in Table
\ref{tab:style}, confirm that the style of the resulting text can be
successfully modulated to the desired degree and that the quality
and diversity are only slightly degraded at moderate increases of
the probability of adjectival clause generation.

\subsection{Human Evaluation} \label{sec:humanevalresults}

In order to better assess the quality of the generated text,
we also include a human evaluation. For this, we took a sample
of 60 sentences of each model under study, including
also a sample of the same size from the validation data,
to serve as reference. The sentences were evaluated
by a pool of annotators,
who were requested to rate the sentence in an integer scale
from 1 to 5, taking into account its fluency and correctness.

The pack of sentences rated by each annotator contained 10
sentences from each of the models under evaluation.
Each sentence under evaluation was part of the packs of
3 evaluators; this redundancy was used to measure the discrepancies
in the rating of each sentence
among annotators, which was quantified by means of the
average per-sentence standard deviation.

\begin{table}[ht]
\centering
\fontsize{10pt}{14pt}\selectfont
\setlength{\tabcolsep}{3.5pt}
\begin{tabular}{r | c c }
\multicolumn{1}{c}{\multirow{2}{*}{\textbf{Model}}}  & \textbf{Average}
 & \textbf{Per sentence}  \\
 & \textbf{rating} & \textbf{avg. stddev} \\
\hline
AWD-LSTM (w) & 3.08 & 0.74\\
Transformer (w) & 3.43 & 0.78 \\
\textsc{ItExp} (w) & 3.28 & 0.73 \\
\hline
AWD-LSTM (sw) & 2.66 & 0.68 \\
Transformer (sw) & 3.33 & 0.83 \\
\textsc{ItExp} (sw) & 3.09 & 0.70 \\
\hline
Valid sample & 4.49 & 0.61 \\
\end{tabular}
\caption{\label{tab:human} Human evaluation for the different models.}
\end{table}

Table \ref{tab:human} shows the statistics of the obtained 
ratings, were we can see the average rating of the sentences
generated by each model, together with the average per-sentence
standard deviation, to understand how different the ratings for
each sentence were among the different evaluator ratings.
We can see that the highest human ratings were obtained by
the Transformer, both with word and subword-level vocabularies,
followed by \textsc{ItExp} and then AWD-LSTM.

\begin{table}[ht]
\centering
\fontsize{10pt}{14pt}\selectfont
\setlength{\tabcolsep}{3.5pt}
\begin{tabular}{r | c c }
\textbf{Adjective}  & \textbf{Average}
 & \textbf{Per sentence}  \\
\textbf{probability} & \textbf{rating} & \textbf{avg. stddev} \\
\hline
$\times 1$ & 3.28 & 0.73 \\
$\times 10$ & 3.16 & 0.79 \\
$\times 20$ & 2.98 & 0.84 \\
$\times 50$ & 3.19 & 0.70 \\
\end{tabular}
\caption{\label{tab:humanstyle} Human evaluation for
\textsc{ItExp} (w) models with increased
adjectival construction probability.}
\end{table}

Table \ref{tab:humanstyle} shows the human evaluation for the
models from the style variation experiments presented in
Table \ref{tab:style}. As we can see, there is a small degradation
in quality as we force high levels of adjectival presence.

\section{Further Comparison with Real Text}

\begin{figure}[ht]
\centering
\includegraphics[width=.95\linewidth]{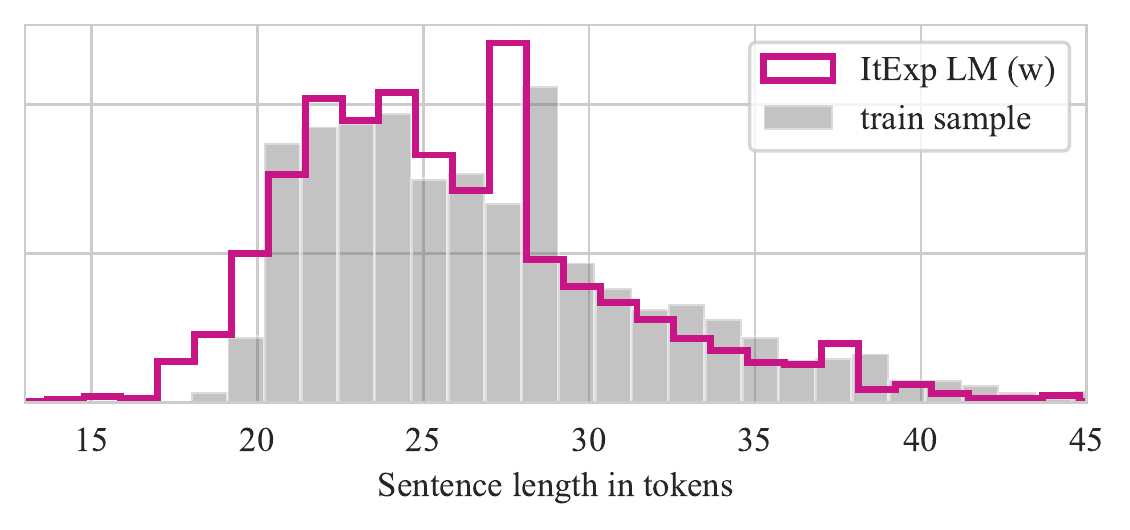}
\caption{Distribution of generated text length.
\label{fig:lengthhistw}}
\end{figure}

Given that the generation process in iterative expansion LMs is
not sequential, we studied the distribution of the sentence lengths
it generates. This is shown in Figure \ref{fig:lengthhistw} for
the text generated by a word-level iterative expansion LM
trained on EMNLP2017 News, along
with the lengths of a sample from the training data.

\begin{figure}[ht]
\centering
\includegraphics[width=.95\linewidth]{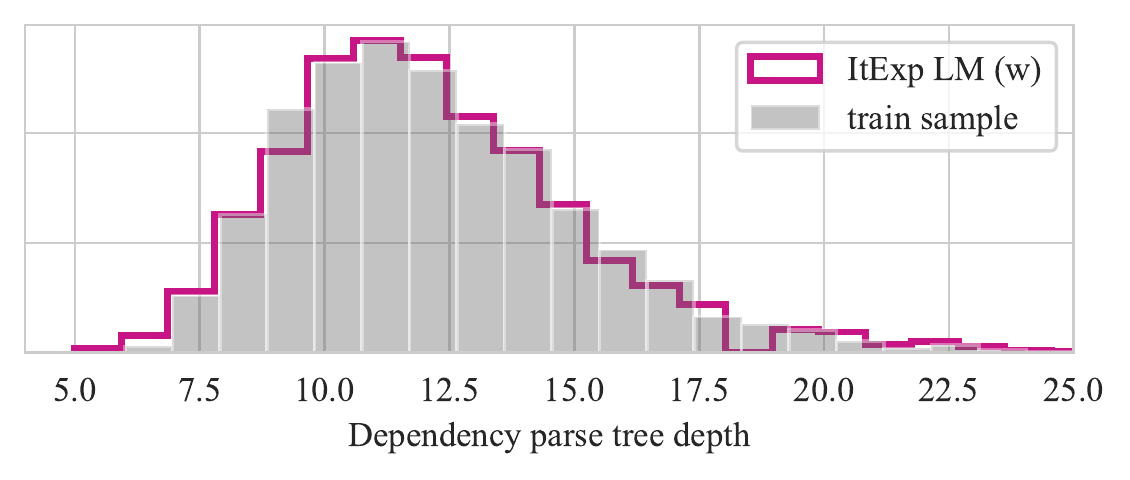}
\caption{Histogram of generated text tree depth.
\label{fig:depthhistw}}
\end{figure}

Iterative expansion LMs generate the dependency parse
tree as they generate text.
We studied the depths of the dependency trees of
generated text in relation to those parsed from the training
data, as shown in Figure \ref{fig:depthhistw}.

We also measured the degree to which
the generated trees adhere to the trees obtained by parsing
their lexicalized representation.
Specifically, we computed the labeled
and unlabeled attachment scores between both for the text
generated at different softmax temperatures $\tau$.
Attachment scores are the standard performance measure
in dependency parsing and are computed as the percentage
of words that have been assigned the same head as the
reference tree, over a test set. The attachment score is
"labeled" if the dependency label
is taken into account or "unlabeled" otherwise.
As shown in Table \ref{tab:attachmentscores}, the obtained
labeled attachment scores (LAS) and unlabeled attachment
scores (UAS) are very high across the different values of
the generation temperature $\tau$.

\begin{table}[ht]
\centering
\fontsize{10pt}{12pt}\selectfont
\setlength{\tabcolsep}{4pt}
\begin{tabular}{r c c c c c}
\textbf{$\tau$} & $0.7$ & $0.8$ & $0.9$ & $1.0$ & $1.2$ \\
\hline
LAS   & 96.4 & 95.3 & 94.2 & 92.3 & 86.2 \\
UAS & 98.0 & 97.3 & 96.5 & 95.2 & 90.7 \\
\end{tabular}
\caption{\label{tab:attachmentscores} Attachment scores of
the generated trees.}
\end{table}

\begin{table*}[ht]
\fontsize{10pt}{13pt}\selectfont
\setlength{\tabcolsep}{5pt}
\begin{tabular}{p{.95\linewidth}}
\hline
American students were 62 percent more likely to die in a heart attack during the first week of 2004, according to the study.\\
For 150 days, Hillary Clinton will do more to improve access to affordable quality care, support and education funding for millions of Americans, she says.\\
For those on this list, it's likely that I would rather be able to train them up, she said.\\
He made it clear the SNP repeated on Friday as a response, saying they discussed a contract getting the extra cost here.\\
He'll pay \$25, 000 for rent and more buses and bring his collection to The Academy on Channel 31.\\
Six years later, at least eight people died as a result of the shooting.\\
The health prime minister told CNN Thursday that he was willing to back up against the US and remove all of the relevant items at the end of the transition.\\
Then, another man told police that was a friend's friend, and as a child, he made the decision to call his mother.\\
They are 40 - 60 among the top 50, 000 women in the last year in that group since 2014 - 15.\\
They've worked hard on Twitter and they think they've tried to focus on our sport, she said.\\
We like to think that if you try to get this game done, we can get a lower success rate out of 15.\\
\hline
\end{tabular}
\caption{Samples of text generated by iterative expansion LMs with
word vocabulary.
\label{tab:selectedsamplesw}}
\end{table*}

\begin{table*}[ht]
\fontsize{10pt}{13pt}\selectfont
\setlength{\tabcolsep}{5pt}
\begin{tabular}{p{.95\linewidth}}
\hline
I feel that they're going to Syria because we had this explanation, that they have an indication of their advance.\\
The girl's mother told the group of three she needed treatment and the family said her daughter would still be alive with another child.\\
But she added: "The data is important to the EU that the UK can attract more businesses.\\
Though he also spoke to Mr Wilson on Saturday morning at the Netherlands Police trial, Johnson referred it to the No. 1 commission.\\
It's a collective belief and it's a statement to us, he said.\\
It's just the first thing we're feeling now and I don't like it.\\
So if you want to be sitting in a garden, you have to wait for something to make sure that this does not end.\\
So, for example, we need to argue about what the president did, but I'm just interested in having any talk.\\
The British defence ministry confirmed action had been taken at the hospital but could not confirm the details until now.\\
We'll ask for a fair share of Russia to stop border security, particularly for people of color, he added.\\
\hline
\end{tabular}
\caption{Samples of text generated by iterative expansion LMs
with subword vocabulary.
\label{tab:selectedsamplessw}}
\end{table*}

\subsection{Quantification of the Generation Speedup}
\label{sec:speedup}

Text generation with autoregressive models like
LSTM or Transformer models offers a linear computational complexity with
respect to the length of the generated sequence.
In comparison, the dependency tree-driven decoding used
by iterative expansion LMs generates text in parallel for
each branch in the tree. If the tree was a perfectly balanced
binary tree, then the computational complexity would be logarithmic.
However, dependency trees in general are not balanced and,
given the tree binarization postprocessing that we introduce, the
parallelization is slightly reduced.

\begin{figure}[ht!]
\centering
\includegraphics[width=.95\linewidth]{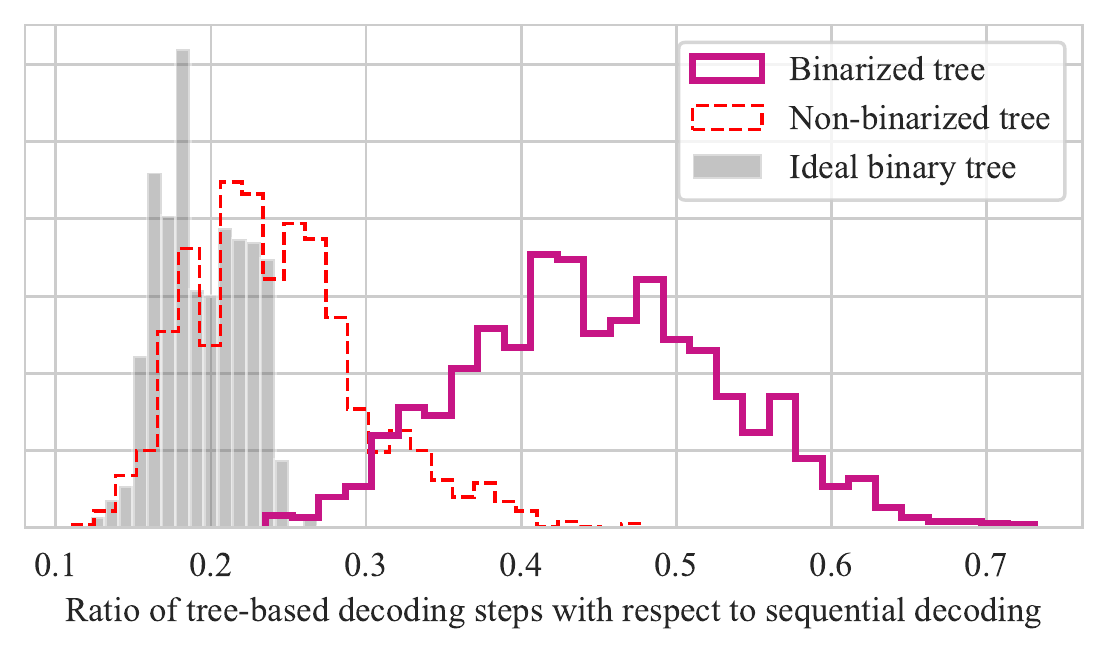}
\caption{Histogram of the ratio of the decoding steps needed to generate
a sentence with tree-based decoding with respect to
sequential generation.
\label{fig:depthlengthratio}}
\end{figure}

Figure \ref{fig:depthlengthratio} shows the speedup of
the needed decoding steps of tree-based decoding with respect
of auto-regressive decoding, taking a sample of the training
data and computing the needed steps to decode them should the
sentences have an idealized binary dependency parse tree, a
normal parse tree, and a binarized parse tree. On average,
the binarized parse tree, which is the decoding
used by iterative expansion LMS, needs only 45\% of the decoding steps
needed by autoregressive decoding.

\subsection{Generation Examples}

Table \ref{tab:selectedsamplesw} 
shows a selection of text samples generated by iterative expansion
LMs with a word-level vocabulary, while
Table \ref{tab:selectedsamplessw} shows samples generated
with a subword-level vocabulary.
We can see that, despite being generated non-sequentially and
each branch of the dependency parse tree being generated in parallel,
the resulting sentences maintain coherence and syntactic
agreement, confirming that conditioning on the token
dependencies in the parse tree provides enough information
to generate it while speeding up the decoding process.

\section{Conclusion} \label{sec:conclusion}

In this work, we presented iterative expansion LMs, which are
iterative
non-autoregressive text generation models that rely on
syntactic dependency trees to generate sentence tokens in parallel.
As opposed to other syntax-driven generation mechanisms, 
the training of iterative expansion LMs can be naturally computed
in batches and they are amenable to subword-level vocabularies.

We showed that our proposed method generates text with quality
between LSTMs and Transformers, with comparable diversity,
both regarding automatic measurements and human judgement,
while generating text in half of the decoding steps needed by
sequential LMs, and also allowing direct control over the generation
process at the syntactic
level, enabling the induction of stylistic variations in
the generated text.

Our code is available as open source at 
\url{https://github.com/noe/iterative_expansion_lms} .

\ifaclfinal

\section*{Acknowledgments}

This work is partially supported
by Lucy Software / United Language Group (ULG)
and the Catalan Agency for Management of University and Research
Grants (AGAUR) through an Industrial Ph.D. Grant.
This work also is supported in part by the Spanish
Ministerio de Economía y Competitividad, the European Regional
Development Fund through the postdoctoral senior grant
Ramón y Cajal and by the Agencia Estatal de Investigación
through the projects EUR2019-103819, PCIN-2017-079 and PID2019-107579RB-I00 / AEI / 10.13039/501100011033

\fi

\bibliography{emnlp2020}
\bibliographystyle{acl_natbib}

\clearpage

\appendix

\section*{Supplementary Material: Appendix}

\section{Dataset Statistics} \label{sec:datasetstatistics}

Table \ref{tab:datasetsize} summarizes the statistics of the
EMNLP2017 News dataset used in our experiments. The
training/validation/test split was taken from the work
by \citet{holtzman2019curious}.%
\footnote{The EMNLP 2017 News data can be downloaded from
\url{https://github.com/pclucas14/GansFallingShort/tree/master/real_data_experiments/data/news}}

\begin{table}[ht]
\centering
\fontsize{10pt}{12pt}\selectfont
\begin{tabular}{r c c c }
 & \textbf{train} & \textbf{valid} & \textbf{test} \\
\hline
sentences & 268k & 10k & 10k \\
iterations & 3.2M & 122k & 122k \\
expansion vocab & 904 \\
terminal vocab  & 8195 \\
\end{tabular}
\caption{\label{tab:datasetsize} Statistics of the EMNLP2017 News dataset.}
\end{table}

\section{Data Processing Details} \label{sec:preprocessing}

\textbf{EMNLP2017 News dataset}.
The tokenization of the EMNLP2017 News dataset is very nonstandard.
To appropriately prepare it to be used as input to
the syntactic annotation tool \texttt{corenlp}, we
detokenized the text and then retokenized it again with the Moses
tokenizer. For the experiments with BPE, we created the
subword vocabulary with 4000 merge operations and without
further constraining the size of the resulting vocabulary.

\textbf{Text generation with AWD-LSTM}. AWD-LSTM
is trained with ``continuous'' text batches. This implies
that when used for text generation, it likewise generates text.
To obtain a predetermined number
of sentences, we used AWD-LSTM to generate a fixed number
of tokens (e.g., 200). Then, we split this text at the
\texttt{<eos>} boundaries and removed the first and last
sentences to avoid incomplete ones. We repeated
this procedure until we had the target number of
sentences.

\textbf{Text generation with the Transformer}.
A Transformer LM was trained following the data
preparation instructions in the fairseq examples.%
\footnote{\url{https://github.com/pytorch/fairseq/tree/master/examples/language_model}}

\textbf{Quality vs. diversity plots}. The generated text
was un-BPE'ed (for the subword-level models) and detokenized
by means of the Moses \texttt{detokenizer.perl} script.
Then, it was tokenized with
the Moses \texttt{tokenizer.perl} script, and the BLEU scores
were computed with the NLTK \texttt{corpus\_bleu}
function \cite{loper02nltk}, without smoothing.

\textbf{GPT-2 perplexity computation}. The text that served
as input to GPT-2 was properly detokenized before
applying the model's own BPE tokenization.

\section{Computing Infrastructure}

All the experiments presented in this work were trained
using 4 nvidia 1080Ti GPUs. The exact number of GPUs used
for each experiment depended on
the total batch size (which is reported in
the hyperparameter tables on the next section) and
the available memory in each GPU.

\section{Hyperparameters} \label{sec:hyperparams}

In this section, we present the detailed hyperparameters
used in the experiments presented in this work. They
were obtained by manual exploration, observing the
behavior of the loss over the training and validation
sets of each dataset. The number of manual
hyperparameter search trials were less than 10
for each model.

The hyperparameters of the iterative expansion LM models
used for the text generation experiments presented in Figure 
\ref{fig:bleuvsselfbleu5},
for both the word and subword vocabulary variants, are
shown in Table \ref{tab:textgenhyperparams}.

\begin{table}[ht]
\centering
\fontsize{10pt}{12pt}\selectfont
\begin{tabular}{r c}
\hline
num. layers & 6\\
num. heads  & 8\\
embed. size & 1024\\
batch size & 16384 \\
num. params     & 96M \\
\hline
\end{tabular}
\caption{\label{tab:textgenhyperparams} Hyperparameters
of the iterative expansion LM used in the 
text generation experiments.}
\end{table}

The hyperparameters of the AWD-LSTM baseline are presented
in Table \ref{tab:textgenhyperparamslstm}.
Note that the
AWD-LSTM variant used as a baseline is the base LM without
the continuous cache pointer mechanism, with tied weights.
Additionally, note that the terminal and expansion vocabulary sizes
are different, which leads to a different
size of the expansion embedding table and therefore
to a different total number of parameters for the same
values of the rest of the hyperparameters.

\begin{table}[ht]
\centering
\fontsize{10pt}{12pt}\selectfont
\begin{tabular}{r c}
\hline
hidden size & 1150\\
embed. size & 400\\
num. layers & 3\\
batch size & 20\\
BPTT       & 70\\
num. params     & 23.5M\\
\hline
\end{tabular}
\caption{\label{tab:textgenhyperparamslstm}
AWD-LSTM baseline hyperparameters.}
\end{table}

The hyperparameters of the Transformer baseline are presented
in Table \ref{tab:textgenhyperparamstransformer}. We used the
implementation of the fairseq library and tuned it on the
training and validation data.

\begin{table}[ht]
\centering
\fontsize{10pt}{12pt}\selectfont
\begin{tabular}{r c}
\hline
num. layers & 6\\
num. heads  & 4\\
embed. size & 512\\
batch size & 16384 \\
num. params     & 17M \\
\hline
\end{tabular}
\caption{\label{tab:textgenhyperparamstransformer}
Transformer baseline hyperparameters.}
\end{table}

The batch size for \textsc{ItExp} is expressed in total number
of tokens, while for AWD-LSTM it is expressed as number
of sentences, which, when multiplied by the back-propagation through
time (BPTT) length, gives the total number of tokens per batch.
Note that the criteria for the optimum batch size
differ for transformers and LSTMs.

To sample from both our proposed model and the baselines, we
use nucleus sampling \cite{holtzman2019curious} with $p=0.9$.

\section{Generated Text Samples} \label{sec:generationsamples}

\subsection{Text generated at different values of $\tau$}

Table \ref{tab:textsamples} presents sentences generated by
iterative expansion LMs trained on EMNLP2017 News at different values
of the final softmax temperature $\tau$. They have not been
cherry-picked.

\subsection{Style Variation Samples}

Table \ref{tab:adjsamples} show samples of sentences generated
with an altered probability of generating adjectival constructions
that is ten times higher, which are not cherry-picked.

\begin{table*}[ht]
\fontsize{9.5pt}{12pt}\selectfont
\setlength{\tabcolsep}{5pt}
\begin{tabular}{p{10mm} p{143mm}}
\hline
\multirow{7}{*}{$\tau=0.7$} &
We're really looking forward to seeing the world in a positive way from the main stage in my life, \textquotedbl{}he said.\\
& I will do everything to make me feel comfortable with myself, and tell you that I can go out and play my part.\\
& I think I'm one of the first people I think we need the people to make the most of it.\\
& The company said she would go on TV and was concerned for the welfare of hundreds of thousands of customers.\\
& \textquotedbl{}It was a one - child policy by one,\textquotedbl{} Mr. Trump told the Financial Times.\\
\hline
\multirow{9}{*}{$\tau=0.8$} &
The good news is that the government is likely to build a wall between the country's population and younger voters.\\
& We can't imagine the figures will increase our interest rates in December and December, with a cost of around £2 billion.\\
& We feel it feels as if this was the result of someone acting in life - threatening, and it made sense.\\
& I like the president - elect, I would want to play fair, and I want someone who is more conservative than that.\\
& We have to show that we have the sort of thing we need as we want of doing what we do.\\
& He also sent out a letter to Tony Abbott, who asked him for a response to Russia's intervention in Ukraine.\\
\hline
\multirow{10}{*}{$\tau=0.9$} &
She said she encouraged her husband to start the company \textquotedbl{}to fight State and Qaeda,\textquotedbl{} and that he would send them to Iraq.\\
& Clinton's appeal means that Bill Clinton on Monday is limited to the amount of the national education budget for the Democrats.\\
& But the weak drinks industry may leave an impression, that key cash restrictions would be a disaster for your business of car.\\
& When Hillary Clinton reporters considered the moment after the election he would bring out their criticism of women for the attacks \textquotedbl{}black identity.\\
& A Home Office spokesperson said: \textquotedbl{}We are aware of the game and are travelling as people are far away from Europe.\\
\hline
\multirow{8}{*}{$\tau=1.0$} &
He produced a decent player, and became the fifth player in the eighth game, and helped Williams to his rally to Miami. \\
& George Osborne, which exposed Labor last month, was reportedly referring to Mr Trump's launch by a senior campaign policy official on Jan.\\
& Almost 60 per cent of them believe it was the first time they had joined the coalition to promote civil war and human rights.\\
& We won't get anywhere, so we had to make that decision and it was a present and say, \textquotedbl{}Is it?\\
& I see what happens, I'm just trying to do something this way, and I don't want\\
\hline
\multirow{9}{*}{$\tau=1.1$} &
The Post Office continues to track hard motion, some local policy experts say, attacking Donald Trump and Americans in respect and attention throughout Clinton's visit.\\
& At the same time, it came as a tough round debut - that had a good game in 60 years.\\
& President Obama has made a huge gains outside from his historic trip to the U. N. General Assembly in 2009, amid criticism of Congress over criticism of Texas Sen. Ted Cruz and Florida Sen. Marco Rubio focusing on Ben Carson.\\
& And it's still possible for us to make a change for a platform; At the moment, it makes the name of the planet.\\
& His return to Poland was a blow to the EU's second - biggest market, which has the option of waiting view longer in the way of a EU visa.\\
\hline
\multirow{9}{*}{$\tau=1.2$} &
I tell our friends to write stories about their mixed ways: can you ask if something is obvious again?\\
& Researchers also noted that jobs' growth assets fear UK taxpayers will forget if real estate wages and a free living wage could be affected by the plan.\\
& \textquotedbl{}All on the street, players and events are speaking up with other teams because they are tired that we should have stuck faster, we don't agree with how our players looks, so you really enjoy playing more,\textquotedbl{} he said.\\
& Labour were eventually advised over a quiet situation within two groups \textquotedbl{}eat and exercise with speed at all, however, say.\\
& The result may be to leave the house in 12 seasons or complete with a personal outdoor work between 6 - year - old.\\
\hline
\end{tabular}
\caption{Samples of text generated by iterative expansion LMs (w) for
different softmax temperatures (not cherry-picked).
\label{tab:textsamples}}
\end{table*}

\begin{table*}[ht]
\fontsize{10pt}{12pt}\selectfont
\setlength{\tabcolsep}{5pt}
\begin{tabular}{p{153mm}}
\hline
\textquotedbl{}The last judge appeal is to focus on the single many main causes of attempted murder,\textquotedbl{} he said.\\
\textquotedbl{}I ask if you are willing to it to say yes and have a serious conversation about the way that I've been prepared,\textquotedbl{} he said.\\
I can just make improvements we need to keep this message going, and we cannot believe that we treat the British Labour badly.\\
I had guys created, and I couldn't see that stronger, and I thought they could do, but it turned out.\\
The same poll leaves 75\% of the voters vote and 48 points in 2012, a standard national measure released last month.\\
\hline
\end{tabular}
\caption{Samples of style variation with adj. probability$\times 10$ from Table
\ref{tab:style} (not cherry-picked).
\label{tab:adjsamples}}
\end{table*}

\subsection{Iterative Expansion Intermediate States}

Figure \ref{fig:iterationsample} shows the full
generation process of iterative expansion LMs for
some sample sentences.

\begin{figure*}[ht]
\fontsize{6.5pt}{12pt}\selectfont
\setlength{\tabcolsep}{0.5pt}
\begin{tabularx}{\linewidth}{p{6mm} c}
\multicolumn{1}{l}{Iteration 1}\\
\hline
\inputtok: & \texttt{[ROOT]}\\
\outputtok: & failure\\
\outputexp: & \texttt{[nsubj-HEAD-punct]}\\
\end{tabularx}
\begin{tabularx}{\linewidth}{p{6mm} c c c}
\multicolumn{3}{l}{Iteration 2}\\
\hline
\inputtok: & \texttt{[nsubj]} & failure & \texttt{[punct]}\\
\outputtok: & It & \texttt{[pad]} & ,\\
\outputexp: & \texttt{[HEAD-cop]} & \texttt{[pad]} & \texttt{[HEAD-cc]}\\
\end{tabularx}
\begin{tabularx}{\linewidth}{p{6mm} c c c c c}
\multicolumn{5}{l}{Iteration 3}\\
\hline
\inputtok: & It & \texttt{[cop]} & failure & , & \texttt{[cc]}\\
\outputtok: & \texttt{[pad]} & was & \texttt{[pad]} & \texttt{[pad]} & and\\
\outputexp: & \texttt{[pad]} & \texttt{[HEAD-det]} & \texttt{[pad]} & \texttt{[pad]} & \texttt{[HEAD-conj]}\\
\end{tabularx}
\begin{tabularx}{\linewidth}{p{6mm} c c c c c c c}
\multicolumn{7}{l}{Iteration 4}\\
\hline
\inputtok: & It & was & \texttt{[det]} & failure & , & and & \texttt{[conj]}\\
\outputtok: & \texttt{[pad]} & \texttt{[pad]} & a & \texttt{[pad]} & \texttt{[pad]} & \texttt{[pad]} & knew\\
\outputexp: & \texttt{[pad]} & \texttt{[pad]} & \texttt{[HEAD]} & \texttt{[pad]} & \texttt{[pad]} & \texttt{[pad]} & \texttt{[nsubj-HEAD-ccomp]}\\
\end{tabularx}
\begin{tabularx}{\linewidth}{p{6mm} c c c c c c c c c}
\multicolumn{9}{l}{Iteration 5}\\
\hline
\inputtok: & It & was & a & failure & , & and & \texttt{[nsubj]} & knew & \texttt{[ccomp]}\\
\outputtok: & \texttt{[pad]} & \texttt{[pad]} & \texttt{[pad]} & \texttt{[pad]} & \texttt{[pad]} & \texttt{[pad]} & we & \texttt{[pad]} & be\\
\outputexp: & \texttt{[pad]} & \texttt{[pad]} & \texttt{[pad]} & \texttt{[pad]} & \texttt{[pad]} & \texttt{[pad]} & \texttt{[HEAD]} & \texttt{[pad]} & \texttt{[advmod-HEAD-punct]}\\
\end{tabularx}
\begin{tabularx}{\linewidth}{p{6mm} c c c c c c c c c c c}
\multicolumn{11}{l}{Iteration 6}\\
\hline
\inputtok: & It & was & a & failure & , & and & we & knew & \texttt{[advmod]} & be & \texttt{[punct]}\\
\outputtok: & \texttt{[pad]} & \texttt{[pad]} & \texttt{[pad]} & \texttt{[pad]} & \texttt{[pad]} & \texttt{[pad]} & \texttt{[pad]} & \texttt{[pad]} & far & \texttt{[pad]} & ,\\
\outputexp: & \texttt{[pad]} & \texttt{[pad]} & \texttt{[pad]} & \texttt{[pad]} & \texttt{[pad]} & \texttt{[pad]} & \texttt{[pad]} & \texttt{[pad]} & \texttt{[advmod-HEAD-nsubj]} & \texttt{[pad]} & \texttt{[HEAD-dep]}\\
\end{tabularx}
\begin{tabularx}{\linewidth}{p{6mm} c c c c c c c c c c c c c c}
\multicolumn{14}{l}{Iteration 7}\\
\hline
\inputtok: & It & was & a & failure & , & and & we & knew & \texttt{[advmod]} & far & \texttt{[nsubj]} & be & , & \texttt{[dep]}\\
\outputtok: & \texttt{[pad]} & \texttt{[pad]} & \texttt{[pad]} & \texttt{[pad]} & \texttt{[pad]} & \texttt{[pad]} & \texttt{[pad]} & \texttt{[pad]} & how & \texttt{[pad]} & ball & \texttt{[pad]} & \texttt{[pad]} & so\\
\outputexp: & \texttt{[pad]} & \texttt{[pad]} & \texttt{[pad]} & \texttt{[pad]} & \texttt{[pad]} & \texttt{[pad]} & \texttt{[pad]} & \texttt{[pad]} & \texttt{[HEAD]} & \texttt{[pad]} & \texttt{[det-HEAD-aux]} & \texttt{[pad]} & \texttt{[pad]} & \texttt{[HEAD-parataxis]}\\
\end{tabularx}
\begin{tabularx}{\linewidth}{p{6mm} c c c c c c c c c c c c c c c c c}
\multicolumn{17}{l}{Iteration 8}\\
\hline
\inputtok: & It & was & a & failure & , & and & we & knew & how & far & \texttt{[det]} & ball & \texttt{[aux]} & be & , & so & \texttt{[parataxis]}\\
\outputtok: & \texttt{[pad]} & \texttt{[pad]} & \texttt{[pad]} & \texttt{[pad]} & \texttt{[pad]} & \texttt{[pad]} & \texttt{[pad]} & \texttt{[pad]} & \texttt{[pad]} & \texttt{[pad]} & the & \texttt{[pad]} & would & \texttt{[pad]} & \texttt{[pad]} & \texttt{[pad]} & have\\
\outputexp: & \texttt{[pad]} & \texttt{[pad]} & \texttt{[pad]} & \texttt{[pad]} & \texttt{[pad]} & \texttt{[pad]} & \texttt{[pad]} & \texttt{[pad]} & \texttt{[pad]} & \texttt{[pad]} & \texttt{[HEAD]} & \texttt{[pad]} & \texttt{[HEAD]} & \texttt{[pad]} & \texttt{[pad]} & \texttt{[pad]} & \texttt{[nsubj-HEAD-xcomp]}\\
\end{tabularx}
\begin{tabularx}{\linewidth}{p{6mm} c c c c c c c c c c c c c c c c c c c}
\multicolumn{19}{l}{Iteration 9}\\
\hline
\inputtok: & It & was & a & failure & , & and & we & knew & how & far & the & ball & would & be & , & so & \texttt{[nsubj]} & have & \texttt{[xcomp]}\\
\outputtok: & \texttt{[pad]} & \texttt{[pad]} & \texttt{[pad]} & \texttt{[pad]} & \texttt{[pad]} & \texttt{[pad]} & \texttt{[pad]} & \texttt{[pad]} & \texttt{[pad]} & \texttt{[pad]} & \texttt{[pad]} & \texttt{[pad]} & \texttt{[pad]} & \texttt{[pad]} & \texttt{[pad]} & \texttt{[pad]} & you & \texttt{[pad]} & wait\\
\outputexp: & \texttt{[pad]} & \texttt{[pad]} & \texttt{[pad]} & \texttt{[pad]} & \texttt{[pad]} & \texttt{[pad]} & \texttt{[pad]} & \texttt{[pad]} & \texttt{[pad]} & \texttt{[pad]} & \texttt{[pad]} & \texttt{[pad]} & \texttt{[pad]} & \texttt{[pad]} & \texttt{[pad]} & \texttt{[pad]} & \texttt{[HEAD]} & \texttt{[pad]} & \texttt{[mark-HEAD-punct]}\\
\end{tabularx}
\begin{tabularx}{\linewidth}{p{6mm} c c c c c c c c c c c c c c c c c c c c c}
\multicolumn{21}{l}{Iteration 10}\\
\hline
\inputtok: & It & was & a & failure & , & and & we & knew & how & far & the & ball & would & be & , & so & you & have & \texttt{[mark]} & wait & \texttt{[punct]}\\
\outputtok: & \texttt{[pad]} & \texttt{[pad]} & \texttt{[pad]} & \texttt{[pad]} & \texttt{[pad]} & \texttt{[pad]} & \texttt{[pad]} & \texttt{[pad]} & \texttt{[pad]} & \texttt{[pad]} & \texttt{[pad]} & \texttt{[pad]} & \texttt{[pad]} & \texttt{[pad]} & \texttt{[pad]} & \texttt{[pad]} & \texttt{[pad]} & \texttt{[pad]} & to & \texttt{[pad]} & .\\
\outputexp: & \texttt{[pad]} & \texttt{[pad]} & \texttt{[pad]} & \texttt{[pad]} & \texttt{[pad]} & \texttt{[pad]} & \texttt{[pad]} & \texttt{[pad]} & \texttt{[pad]} & \texttt{[pad]} & \texttt{[pad]} & \texttt{[pad]} & \texttt{[pad]} & \texttt{[pad]} & \texttt{[pad]} & \texttt{[pad]} & \texttt{[pad]} & \texttt{[pad]} & \texttt{[HEAD]} & \texttt{[pad]} & \texttt{[HEAD]}\\
\end{tabularx}
\caption{Generation of sentence ``It was a failure, and
we knew how far the ball would be, so you have to wait.''
\label{fig:iterationsample}}
\end{figure*}

\subsection{Iterative Expansion Generated Trees}

Figure \ref{fig:treesamples} shows examples of generated
sentences together with their dependency trees.

\begin{sidewaysfigure*}[ht]
\begin{dependency}[font=\footnotesize,text only label, label style={above}]]
\tikzstyle{word}=[font=\small]
\begin{deptext}[column sep=.2cm,ampersand replacement=\^]
|[word]| After \^ |[word]| the \^ |[word]| Olympics \^ |[word]| I \^ |[word]| was \^ |[word]| able \^ |[word]| to \^ |[word]| get \^ |[word]| well \^ |[word]| and \^ |[word]| play \^ |[word]| in \^ |[word]| the \^ |[word]| Champions \^ |[word]| League \^ |[word]| , \^ |[word]| and \^ |[word]| I \^ |[word]| believe \^ |[word]| I \^ |[word]| can \^ |[word]| do \^ |[word]| and \^ |[word]| will \^ |[word]| be \^ |[word]| ready \^ |[word]| for \^ |[word]| it \^ |[word]| . \\
\end{deptext}
\depedge{3}{1}{case}
\depedge{1}{2}{det}
\depedge[edge height=1.5cm]{6}{3}{nmod}
\depedge{3}{4}{nsubj}
\depedge{4}{5}{cop}
\deproot[edge height=3cm]{6}{ROOT}
\depedge{8}{7}{mark}
\depedge{6}{8}{xcomp}
\depedge{8}{9}{advmod}
\depedge{9}{10}{cc}
\depedge{10}{11}{conj}
\depedge[edge height=1.5cm]{15}{12}{case}
\depedge{12}{13}{det}
\depedge{13}{14}{compound}
\depedge[edge height=2.3cm]{11}{15}{nmod}
\depedge{15}{16}{punct}
\depedge{16}{17}{cc}
\depedge{19}{18}{nsubj}
\depedge{17}{19}{conj}
\depedge{22}{20}{nsubj}
\depedge{20}{21}{aux}
\depedge{19}{22}{ccomp}
\depedge{22}{23}{cc}
\depedge{26}{24}{aux}
\depedge{24}{25}{cop}
\depedge{23}{26}{conj}
\depedge{28}{27}{case}
\depedge{26}{28}{nmod}
\depedge{28}{29}{punct}
\end{dependency}
\vspace{.5cm}
\begin{dependency}[font=\footnotesize,text only label, label style={above}]]
\tikzstyle{word}=[font=\small]
\begin{deptext}[column sep=.5cm,ampersand replacement=\^]
|[word]| Six \^ |[word]| years \^ |[word]| later \^ |[word]| , \^ |[word]| at \^ |[word]| least \^ |[word]| eight \^ |[word]| people \^ |[word]| died \^ |[word]| as \^ |[word]| a \^ |[word]| result \^ |[word]| of \^ |[word]| the \^ |[word]| shooting \^ |[word]| . \\
\end{deptext}
\depedge{2}{1}{nummod}
\depedge{3}{2}{nmod:npmod}
\depedge[edge height=2.2cm]{9}{3}{advmod}
\depedge{3}{4}{punct}
\depedge{6}{5}{case}
\depedge{7}{6}{nmod:npmod}
\depedge{8}{7}{nummod}
\depedge[edge height=1.5cm]{4}{8}{nsubj}
\deproot[edge height=3cm]{9}{ROOT}
\depedge{12}{10}{case}
\depedge{10}{11}{det}
\depedge{9}{12}{nmod}
\depedge{15}{13}{case}
\depedge{13}{14}{det}
\depedge{12}{15}{nmod}
\depedge{15}{16}{punct}
\end{dependency}

\begin{dependency}[font=\footnotesize,text only label, label style={above}]
\tikzstyle{word}=[font=\small]
\begin{deptext}[column sep=.2cm,ampersand replacement=\^]
|[word]|" \^ |[word]| I \^ |[word]| can \^ |[word]| always \^ |[word]| do \^ |[word]| it \^ |[word]| with \^ |[word]| these \^ |[word]| powers \^ |[word]| , \^ |[word]| just \^ |[word]| like \^ |[word]| all \^ |[word]| Government \^ |[word]| officials \^ |[word]| here \^ |[word]| in \^ |[word]| Congress \^ |[word]| , \^ |[word]| " \^ |[word]| Trump \^ |[word]| said \^ |[word]| on \^ |[word]| Monday \^ |[word]| . \\
\end{deptext}
\depedge[edge height=1.5cm]{5}{1}{punct}
\depedge{1}{2}{nsubj}
\depedge{2}{3}{aux}
\depedge{3}{4}{advmod}
\deproot[edge height=3cm]{5}{ROOT}
\depedge{5}{6}{dobj}
\depedge{9}{7}{case}
\depedge{7}{8}{det}
\depedge{6}{9}{nmod}
\depedge{9}{10}{punct}
\depedge[edge height=2.5cm]{22}{11}{advmod}
\depedge{11}{12}{mark}
\depedge{15}{13}{det}
\depedge{13}{14}{compound}
\depedge{12}{15}{nsubj}
\depedge{15}{16}{advmod}
\depedge{18}{17}{case}
\depedge{16}{18}{nmod}
\depedge{18}{19}{punct}
\depedge{19}{20}{punct}
\depedge{20}{21}{nsubj}
\depedge[edge height=3.3cm]{10}{22}{advcl}
\depedge{24}{23}{case}
\depedge{22}{24}{nmod}
\depedge{24}{25}{punct}
\end{dependency}
\vspace{1cm}
\caption{Examples of generated sentences, together with
their generated parse trees.}
\label{fig:treesamples}
\end{sidewaysfigure*}

\end{document}